\documentclass[letterpaper, 10 pt, conference]{ieeeconf}  

\IEEEoverridecommandlockouts                              

\usepackage[usenames, dvipsnames]{color}
\usepackage{listings}
\usepackage{xcolor}
\usepackage{url}
\usepackage{float}
\usepackage{amsmath,amssymb,amsthm}
\usepackage{mathtools} 
\usepackage[english]{babel}

\usepackage{epsfig}
\usepackage{epstopdf}
\usepackage{subcaption}
\usepackage{multirow}
\usepackage{graphicx}

\usepackage[export]{adjustbox}
\usepackage[boxruled,vlined,linesnumbered]{algorithm2e}

\usepackage{hyperref}
\usepackage[autostyle, english = american]{csquotes}
\MakeOuterQuote{"}

\title{\LARGE \bf
LES: Locally Exploitative Sampling for \\ Robot Path Planning
}

\author{Sagar Suhas Joshi$^{1}$~ Seth Hutchinson$^{2}$ ~ Panagiotis Tsiotras$^{3}$
	\thanks{$^{1,2,3}$ Institute for Robotics and Intelligent Machines, Georgia Institute of Technology, USA.
		Email: {\small \{sagarsjoshi94, seth, tsiotras\}@gatech.edu}}
}

\begin{document}

\maketitle
\thispagestyle{empty}
\pagestyle{empty}

\begin{abstract}

Sampling-based algorithms solve the path planning problem by generating random samples in the search-space and incrementally growing a connectivity graph or a tree. 
Conventionally, the sampling strategy used in these algorithms is biased towards exploration to acquire information about the search-space. 
In contrast, this work proposes an optimization-based procedure that generates new samples to improve the cost-to-come value of vertices in a neighborhood. 
The application of proposed algorithm adds an exploitative-bias to sampling and results in a faster convergence\footnote{In this work, convergence implies convergence to the optimal solution, unless stated otherwise.} to the optimal solution compared to other  state-of-the-art sampling techniques.
This is demonstrated using benchmarking experiments performed for a variety of higher dimensional robotic planning tasks.
\end{abstract}

\section{INTRODUCTION}
Sampling-based motion planning (SBMP) algorithms have become the default choice for solving robotic planning tasks due to their scalability to higher dimensional problems.
These algorithms do not resort to discretization or explicit construction of the search-space.
Instead, popular single-query SBMP algorithms such as RRT~\cite{lavalle2001randomized} and multi-query algorithms such as PRM~\cite{kavraki1996probabilistic} use a black-box collision checking function to probe a set of random samples and local connections to incrementally build a connectivity graph. 
These algorithms are \textit{probabilistically complete}, i.e., the probability of finding a feasible solution, if it exists, approaches unity as the number of samples tends to infinity.

\textit{Asymptotically optimal} variants of RRT, such as RRT*~\cite{karaman2011sampling}, converge to the optimal solution almost-surely.
These algorithms comprise of two fundamental modules, namely, graph-growth and graph-processing.
The graph-growth module generates random samples, performs nearest neighbor, local steering and collision checking calculations to build a connectivity graph during planning time.
The graph-processing module then tries to improve the cost-to-come value of the vertices by performing operations such as edge rewiring. 
The graph is said to be rewired if the parent of a vertex changes, improving its cost-to-come value.
In particular, the "local rewiring" procedure of RRT* first selects the best parent for a newly initialized vertex. 
It then sees if this new vertex can be a better parent for any of the vertices in its neighborhood. 
The RRT$^{\#}$~\cite{arslan2013use} algorithm provides an extension to the RRT* procedure by "globally rewiring" the graph using dynamic programming.
It uses value-iteration~\cite{arslan2013use} or policy-iteration~\cite{arslan2016incremental} to optimally connect each vertex in the graph in order to minimize their cost-to-come values.
Recently proposed methods such as BIT*~\cite{gammell2015batch} and FMT*~\cite{janson2015fast} also use ideas from dynamic programming and heuristics to obtain faster convergence than RRT*.

Using an intelligent sampling strategy, in conjunction with these graph-processing methods, is effective for accelerating the convergence of SBMP algorithms. 
Uniform random sampling, a widely used approach, biases the graph growth towards vertices with larger Voronoi regions in RRT-style methods \cite{lavalle2001randomized}. 
This results in a rapid exploration of the search-space and is effective for finding an initial solution in single-query scenarios.
However, this strategy like many others, prioritizes acquisition of new information over the improvement of current paths in the planner's graph.
This bias towards exploration can have a detrimental effect on convergence, especially in higher dimensions~\cite{gammell2018informed}.

The algorithm proposed in this work aims to generate new samples that can improve the cost-to-come value of vertices and initiate rewirings.  
This is in contrast to the exploration-biased techniques.
The proposed algorithm first selects a vertex and then generates a new sample in its vicinity.
This sample is generated by solving an optimization problem, wherein the objective is to minimize the sum of cost-to-come value of a vertex and its randomly selected descendants.
The proposed sampling algorithm thus leverages local information to provide an exploitative bias.
The combination of global exploratory and locally exploitative sampling results in faster convergence for SBMP algorithms, as demonstrated by several benchmarking experiments. 
\begin{figure}
    \centering
    \includegraphics[width=0.7\columnwidth]{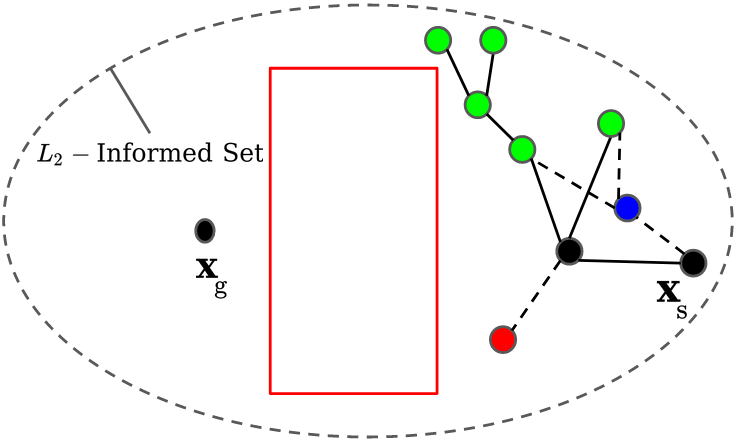}
    \caption{Schematic motivating the proposed LES algorithm, which leverages local information and considers an optimization problem to generate the blue sample. In contrast to the red sample, blue sample can initiate rewirings and improve cost-to-come value of (green) vertices in the graph. }
    \label{fig:les_motivate}
\end{figure}
\section{RELATED WORK}
Many approaches have been suggested to address the exploration-exploitation trade-off in SBMP. 
Akgun and Stilman~\cite{akgun2011sampling} generate samples near a randomly selected state on the current solution path. This local biasing technique increases the probability of improving the current solution at the cost of exploring other homotopy classes.
The RRM algorithm~\cite{alterovitz2011rapidly} adds edges to the current roadmap to balance exploration and refinement.
Techniques such as \cite{urmson2003approaches}, \cite{phillips2004guided}, \cite{persson2014sampling}, \cite{lai2020bayesian}, \cite{rodriguez2006obstacle} use heuristics and obstacle information to guide search during planning.
T-RRT~\cite{jaillet2010sampling} and its variants  \cite{devaurs2013enhancing}, \cite{devaurs2015optimal} implement a transition-test to avoid unhindered exploration in high cost regions.
These approaches provide a way to focus search during planning. 
However, they do not directly address the problem of improving the cost-to-come value of vertices through sampling. 

Unlike the above approaches, Informed Sampling~\cite{gammell2018informed} avoids redundant exploration after an initial solution is discovered. 
It focuses search onto a subset of the search-space, called the Informed Set, that contains all the points that can potentially improve the current solution.
Generating new samples in the Informed Set is thus a necessary (but not sufficient) condition to improve the current solution. 
Relevant Region~\cite{joshi2019relevant}, a subset of the Informed Set, leverages cost-to-come information from the planner's graph to further focus search during planning. 
The combination of Relevant Region and Informed Sampling results in accelerated convergence in uniform and general cost-space environments.
However, these techniques do not generate samples to directly improve the cost-to-come value of vertices.
Hence, some of the samples may fail to trigger any improvement in the planner's graph. 
The sampling algorithm proposed in this work also generates new samples in the Relevant Region to avoid redundant exploration. 
However, it does so by solving an optimization problem aimed towards improving the cost-to-come value of vertices in the graph.
Application of the proposed sampling algorithm thus initiates a higher number of rewirings and results in a faster convergence.
Please see Fig.~\ref{fig:les_motivate} for an illustration of this.

Approaches combining sampling-based planning and local optimizers have also been explored. RABIT*~\cite{choudhury2016regionally} uses CHOMP~\cite{zucker2013chomp} to get feasible, high quality edges connecting any two vertices during a global search performed by BIT*.
However, RABIT* requires pre-computed domain information, such as an obstacle potential function, which may not be available in many practical problems.  
Volumetric Tree*~\cite{kim2019volumetric} addresses this limitation by constructing an approximation of the obstacle-free configuration space on-the-fly. 
However, it relies on uniform random  sampling  for  graph  construction,  which  may  lead to redundant exploration. 
DRRT~\cite{hauer2017deformable} employs a gradient-descent based procedure in the graph-processing module. 
It attempts to optimize the location of vertices to improve their cost-to-come value. 
However, DRRT incurs a higher computational cost due to the extra calls to the nearest-neighbor and collision checking function to ensure edge feasibility after vertex movement.
This work combines ideas from DRRT and \cite{joshi2019relevant} to propose an optimization based sampling procedure.
The proposed method does not require extra calls to the collision checker/nearest-neighbor and can be used in conjunction with any graph-processing module. 

In the following sections, the path planning problem is formally defined, followed by a description and motivation behind the optimization problem to generate new samples.
The proposed sampling algorithm is then discussed and is followed by benchmarking experiments.
\begin{figure}
    \centering
    \includegraphics[width=0.65\columnwidth]{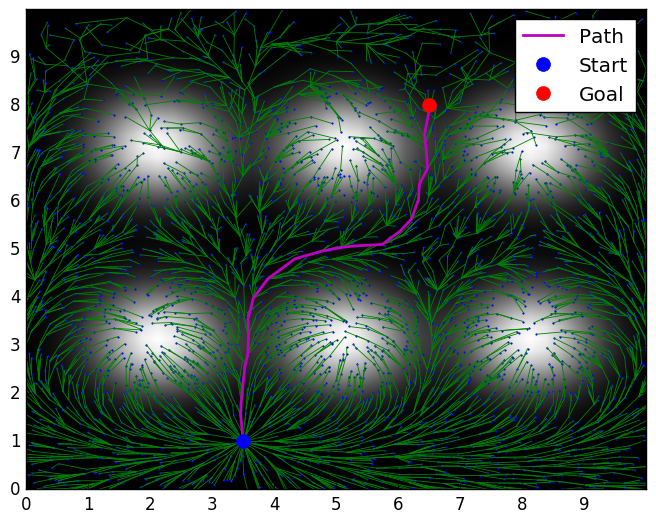}
    \caption{Planning with the proposed LES algorithm on a potential cost-map. The robot incurs a higher cost if it travels in the white regions.}
    \label{fig:potential_costmap}
\end{figure}
\section{PROBLEM DEFINITION}

\subsection{Path Planning Problem}

Consider the search-space $\mathcal{X} \subset \mathbb{R}^d$, with dimension $d$, $d \geq 2$.
Let the obstacle space and free space be denoted by $\mathcal{X}_\mathrm{obs}$ and $\mathcal{X}_\mathrm{free}$ respectively.
Then $\mathcal{X}_\mathrm{free}= \mathrm{cl}(\mathcal{X}\setminus \mathcal{X}_\mathrm{obs})$, where  $\mathrm{cl}(A)$ represents closure of the set $A \subset \mathbb{R}^d$.
Let the cost of moving from a point $\textbf{x}_1 \in \mathcal{X}$ to $\textbf{x}_2 \in \mathcal{X}$ along a path $ \pi:[0,1] \rightarrow \mathcal{X}$, $\pi(0)=\textbf{x}_1$, $\pi(1)=\textbf{x}_2$ be denoted by $\mathrm{c}_{\pi}(\textbf{x}_1,\textbf{x}_2)$,
\begin{equation}
\label{eq:ICcost}
\mathrm{c}_{\pi}(\textbf{x}_1,\textbf{x}_2)= \int_{0}^{1} C(\pi(s)) \  \|\frac{\mathrm{d}\pi(s)}{\mathrm{d}s} \|_2  \ \mathrm{d}s .
\end{equation} 
Here, $C:\mathcal{X} \rightarrow \mathbb{R}_{\geq 0}$ denotes a continuous state cost function.
Note that (\ref{eq:ICcost}) represents the integral of state-cost (IC) metric as a measure of path quality~\cite{jaillet2010sampling}.
The path $\pi$ in (\ref{eq:ICcost}) is assumed to be collision free. The path-cost is infinite otherwise.
The optimal path planning problem can be formally defined as the search for minimum cost path $\pi^*$ from the set of feasible paths $\Pi$ connecting the start state $\textbf{x}_\mathrm{s} \in \mathcal{X}_\mathrm{free}$ to the goal region $\mathcal{X}_\mathrm{goal} \subset \mathcal{X}_\mathrm{free}$,
\begin{equation}
\label{eq:optimalPlanningDef}
\begin{aligned}
\arg \min_{\pi \in \Pi}  & \ \mathrm{c}_{\pi}(\textbf{x}_\mathrm{s},\textbf{x}_\mathrm{g}), \\
\text{subject to:}  & \ \pi(0)= \textbf{x}_\mathrm{s},~ \pi(1)= \textbf{x}_\mathrm{g} \in \mathcal{X}_\mathrm{goal}, \\
& \ \pi(s) \in \mathcal{X}_\mathrm{free}, ~~~ s \in [0,1].  
\end{aligned}
\end{equation}
SBMP algorithms solve the above problem (\ref{eq:optimalPlanningDef}) by constructing a connectivity graph $\mathcal{G}=(V,E)$ with a finite set of vertices $V \subset \mathcal{X}_\mathrm{free}$ and a set of edges $E \subseteq V \times V$. 
The "geometric" versions of SBMP algorithms ignore the kino-dynamic constraints of the robot.
Conventionally, these planners construct an edge $(\textbf{u},\textbf{v}) \in E$ using a straight line path $\pi(s)=\textbf{u}+(\textbf{v}-\textbf{u})s$, $s \in [0,1]$ connecting $\textbf{u}$ and $\textbf{v}$.
Using (\ref{eq:ICcost}), the edge-cost can be denoted as 
\begin{equation}
\label{eq:ICstraightcost}
\mathrm{c}_\ell(\textbf{u},\textbf{v})= \|\textbf{u}-\textbf{v}\|_2 \int_{0}^{1} C(\textbf{u}+(\textbf{v}-\textbf{u})s) \ \mathrm{d}s .
\end{equation}   
SBMP algorithms can perform numerical integration to calculate the edge-cost $\mathrm{c}_\ell(\textbf{u},\textbf{v})$ for any edge $(\textbf{u},\textbf{v}) \in E$.
The graph $\mathcal{G}$ embeds a spanning tree $\mathcal{T}=(V_t,E_t)$ with $V_t=V$ and 
$E_t=\{ (\textbf{u},\textbf{v}) \in E \ | \ \textbf{v}= \mathsf{parent}(\textbf{u})\}$. 
Here, $\mathsf{parent}:V \rightarrow V$ denotes the function mapping a vertex to its unique parent in the tree. 
By definition, we have $\mathsf{parent}(\textbf{x}_\mathrm{s})=\textbf{x}_\mathrm{s}$.
The cost-to-come value $\mathrm{g}_\mathcal{T}(\textbf{v})$ for a vertex $\textbf{v}$ denotes the sum of edge-costs along the path from $\textbf{v}$ to the root $\textbf{x}_\mathrm{s}$ in $\mathcal{T}$. The function $\mathrm{g}_\mathcal{T}:V \rightarrow \mathbb{R}_{\geq 0}$ can be written recursively  as 
\begin{equation}
\label{eq:costTocome}
\mathrm{g}_\mathcal{T}(\textbf{v}) = \mathrm{g}_\mathcal{T}(\textbf{v}_\mathrm{p}) + \mathrm{c}_\ell(\textbf{v}_\mathrm{p},\textbf{v}),
\end{equation}
where $\textbf{v}_\mathrm{p}=\mathsf{parent}(\textbf{v})$. By definition, the recursion ends at $\textbf{x}_\mathrm{s}$ with $\mathrm{g}_\mathcal{T}(\textbf{x}_\mathrm{s})=0$.
Let the set of children for vertex $\textbf{v}$ be denoted by $V_{\textbf{v}}=\{ \textbf{u} \in V \ | \ \textbf{v}= \mathsf{parent}(\textbf{u})\}$ and the number of children by $n_\textbf{v}=|V_{\textbf{v}}|$.
Descendants of a vertex $\textbf{v}$ are all the vertices $\textbf{u} \in V$ whose path from $\textbf{u}$ to the root $\textbf{x}_\mathrm{s}$ in $\mathcal{T}$ contains $\textbf{v}$.
Let $D_{\textbf{v}}$ denote the set of vertices that are descendants of $\textbf{v}$ and  $d_\textbf{v} = |D_{\textbf{v}}|$.
Then,  
\begin{equation}
\label{eq:descendants}
d_\textbf{v} = n_\textbf{v} + \sum_{\textbf{u} \in V_{\textbf{v}}} d_\textbf{u}.
\end{equation}
Note that for a leaf vertex $\textbf{v} \in V$, we have $n_\textbf{v}=d_\textbf{v}=0$.
Let $\mathrm{h}: \mathcal{X} \times \mathcal{X} \rightarrow \mathbb{R}_{\geq 0} $ denote a consistent heuristic function. 
This function obeys the triangle inequality and gives an under-estimate of the path-cost $\mathrm{c}_{\pi}(\textbf{x}_1,\textbf{x}_2)$ between any two points $\textbf{x}_1,\textbf{x}_2 \in \mathcal{X}$. 
An example of function $\mathrm{h}$ is the $L_2$-norm (Euclidean distance).
\begin{figure}
    \centering
    \includegraphics[width=0.8\columnwidth]{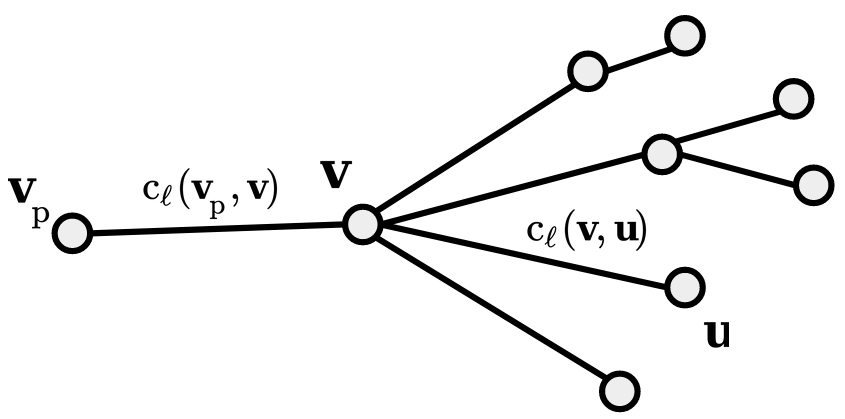}
    \caption{Neighborhood around a vertex $\textbf{v}$. Here, $n_\textbf{v}=4$ and $\widehat{d}_{\textbf{v},V_\textbf{v}}=4+(1+2)=7$.}
    \label{fig:vertex_neighborhood}
\end{figure}
Let $\mathcal{B}^{\epsilon}(\textbf{x}_\mathrm{o})$ denote an $\epsilon$-ball around $\textbf{x}_\mathrm{o} \in \mathcal{X}$, given by 
$\mathcal{B}^{\epsilon}( \textbf{x}_\mathrm{o})=\{ \textbf{x} \in \mathcal{X} \ | \ \|\textbf{x}- \textbf{x}_\mathrm{o}\|_2 < \epsilon \}$, for $\epsilon>0$.
Finally, let $\mu( A )$ denote the Lebesgue measure of the set $A \subset \mathbb{R}^d$.
%
\subsection{Optimization Problem for Sampling}
Given $\mathcal{G}$, the objective of the graph-processing module is to minimize the cost-to-come value of all vertices.
This objective can be written as
\begin{equation}
J_\mathcal{T}  = \sum_{\textbf{u} \in V} \mathrm{g}_\mathcal{T}(\textbf{u}).
\end{equation}
Let $J_\mathcal{T}(\textbf{v})$ denote the terms of $J_\mathcal{T}$ that are dependent only on a particular vertex $\textbf{v} \in V$. 
The position of vertex $\textbf{v}$ impacts the cost-to-come value of itself and its descendants. 
Then, 
\begin{equation}
\label{eq:JTv_1}
J_\mathcal{T}(\textbf{v})=\mathrm{g}_\mathcal{T}(\textbf{v})+\sum_{\textbf{w} \in D_{\textbf{v}}}\mathrm{g}_\mathcal{T}(\textbf{w}).
\end{equation}
Using (\ref{eq:costTocome}). the above equation for $J_\mathcal{T}(\textbf{v})$ can be written in terms of edge-costs $\mathrm{c}_\ell(\textbf{v}_\mathrm{p},\textbf{v})$ and $\mathrm{c}_\ell(\textbf{v},\textbf{u})$.
Here, $\textbf{v}_\mathrm{p}=\mathsf{parent}(\textbf{v})$ and $\textbf{u}$ is any child of $\textbf{v}$.
The edge-cost $\mathrm{c}_\ell(\textbf{v}_\mathrm{p},\textbf{v})$ will appear $1+d_\textbf{v}$ times in total, to calculate the cost-to-come value of $\textbf{v}$ and its descendants. 
Similarly, the edge-cost $\mathrm{c}_\ell(\textbf{v},\textbf{u})$ will appear  $1+d_\textbf{u}$ times in total, to calculate the cost-to-come value of $\textbf{u}$ and its descendants.
Then, 
\begin{equation}
\label{eq:JTv_2}
J_\mathcal{T}(\textbf{v}) =k_1+ (1+d_\textbf{v})\mathrm{c}_\ell(\textbf{v}_\mathrm{p},\textbf{v}) 
+ \sum_{\textbf{u} \in V_{\textbf{v}}} (1+d_\textbf{u}) \mathrm{c}_\ell(\textbf{v},\textbf{u}).
\end{equation}
Note again that equation (\ref{eq:JTv_2}) for $J_\mathcal{T}(\textbf{v})$ only contains terms dependent on $\textbf{v}$.
Other terms are incorporated in the constant $k_1$.
Also, $d_\textbf{v}$ and $d_\textbf{u}$ in (\ref{eq:JTv_2}) are linked by equation (\ref{eq:descendants}). 
A new sample can be generated by first selecting a vertex $\textbf{v}$ and then finding a "better" position for it by optimizing $J_\mathcal{T}$ with respect to $\textbf{v}$.
Note that $\arg\min_{\textbf{v}}J_\mathcal{T}=\arg\min_{\textbf{v}}J_\mathcal{T}(\textbf{v})$.

However, calculating the values of the coefficients $d_\textbf{v}, d_\textbf{u}$ in (\ref{eq:JTv_2}) requires a depth-first search with time complexity of $O(|V_t|)$.
This may get computationally cumbersome, especially as the planner tree grows larger with the number of iterations. 
The vertex data structure in standard implementations of SBMP algorithms (such as OMPL~\cite{sucan2012open}) only stores information about the vertex's children. 
Hence, the following objective function can be considered instead,
\begin{equation}
\begin{aligned}
\label{eq:JTv_hat}
\widehat{J}_{\mathcal{T},V_\textbf{v}}(\textbf{v}) &= k_2+ (1+\widehat{d}_{\textbf{v},V_\textbf{v}})\mathrm{c}_\ell(\textbf{v}_\mathrm{p},\textbf{v}) 
+ \sum_{\textbf{u} \in V_{\textbf{v}}} (1+n_\textbf{u}) \mathrm{c}_\ell(\textbf{v},\textbf{u}),\\
\widehat{d}_{\textbf{v},V_\textbf{v}} & = n_\textbf{v} + \sum_{\textbf{u} \in V_{\textbf{v}}} n_\textbf{u}.
\end{aligned}
\end{equation}
Please see Fig.~\ref{fig:vertex_neighborhood}. 
\begin{figure}
    \centering
    \includegraphics[width=0.9\columnwidth]{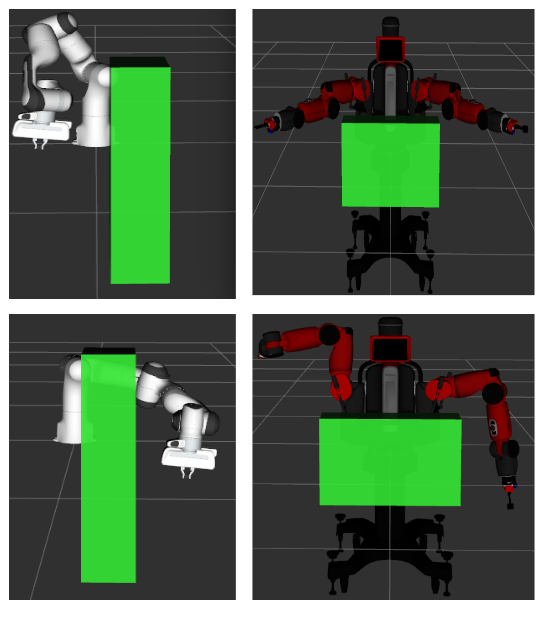}
    \caption{Planning in the joint space of Panda ($\mathbb{R}^7$) and Baxter ($\mathbb{R}^{14}$) manipulator arms. The start and goal positions for both robots are indicated in the top and bottom figures respectively. }
    \label{fig:manipulators}
\end{figure}
Note that minimizing $\widehat{J}_{\mathcal{T},V_\textbf{v}}(\textbf{v})$ in (\ref{eq:JTv_hat}) with respect to $\textbf{v}$ is equivalent to minimizing the cost-to-come values of $\textbf{v}$, the set of children $V_\textbf{v}$ and their children. 
The objective $\widehat{J}_{\mathcal{T},V_\textbf{v}}(\textbf{v})$ can be calculated efficiently with the information contained in the data structure of vertex $\textbf{v}$, without recursing deeper down the tree.
Effectively, $\widehat{J}_{\mathcal{T},V_\textbf{v}}(\textbf{v})$ considers descendants of $\textbf{v}$ upto a depth of $2$. 
This can be generalized to depth-$k$ descendants, at a higher computational cost for calculating the coefficients. 
\IncMargin{.5em}
\begin{algorithm}[t]
	\caption{LES Algorithm Flow}
	\label{alg:LESalgFlow}	
	$V \leftarrow \{ \textbf{x}_\mathrm{s} \} $; $E \leftarrow \phi$; $\mathcal{G} \leftarrow (V,E)$\;	
	\For{$i=1:N$} 
	{
		$c_i \leftarrow \mathsf{getBestSolutionCost()}$\;
		$u_\mathrm{rand} \sim \mathcal{U}(0,1)$\;
		\eIf{$u_\mathrm{rand}<p_\mathrm{LES} \ \normalfont \text{and} \ c_i<\infty$}
		{
			$\textbf{v} \leftarrow \mathsf{chooseVertex}(V_\mathrm{rel}$)\;
			$\hat{\textbf{e}}\leftarrow \mathsf{getGradientDirection}(\textbf{v})$\;
			$\gamma \leftarrow \mathsf{getStepSize}(\textbf{v},\hat{\textbf{e}})$\;
			$\textbf{x}_\mathrm{rand} \leftarrow  \textbf{v}-\gamma\hat{\textbf{e}}$\;
		}
		{
			$\textbf{x}_\mathrm{rand} \leftarrow \mathsf{InformedSampling(c_i)}$
		}
		$\mathsf{Extend}(\textbf{x}_\mathrm{rand})$\;		
		$\mathsf{GraphProcessing}(\mathcal{G})$\;
	}
	\KwRet $\mathcal{G}$
\end{algorithm}
\DecMargin{.5em}

Finally, a random subset of the children, denoted by $\widehat{V_\textbf{v}} \subseteq V_\textbf{v}$, can be selected and a new sample generated by minimizing $\widehat{J}_{\mathcal{T},\widehat{V_\textbf{v}}}(\textbf{v})$. This serves two purposes. First, it  promotes a desirable randomness in the sampling process. 
Second, focusing on the subset $\widehat{V_\textbf{v}}$ effectively assigns a weight of zero for the terms corresponding to the vertices $V_\textbf{v} \setminus \widehat{V_\textbf{v}}$ in the objective (\ref{eq:JTv_hat}).
This can lead to a better improvement in the cost-to-come value of vertices corresponding to $\widehat{V_\textbf{v}}$.
\section{LOCALLY EXPLOITATIVE SAMPLING}
The proposed "Locally Exploitative Sampling (LES)" procedure first selects a vertex $\textbf{v}$ and then generates a new sample considering $\widehat{J}_{\mathcal{T},\widehat{V_\textbf{v}}}(\textbf{v})$.
Expansive Space Trees (EST)~\cite{hsu1997path} and its variants, such as \cite{phillips2004guided}, \cite{persson2014sampling}, also proceed by selecting a vertex and generating a random sample in its vicinity. 
However, the probability of generating a "good" sample (that can improve $\widehat{J}_{\mathcal{T},\widehat{V_\textbf{v}}}(\textbf{v})$) with such random search may decrease rapidly in higher dimensions. 
This is illustrated in the Appendix by considering the problem of minimizing a quadratic function $J_q(\textbf{x})=\textbf{x}^\mathsf{T}\textbf{x}$ with random local search.
The probability of generating a sample that can improve $J_q$ diminishes exponentially with the dimension $d$.
\IncMargin{.5em}
\begin{algorithm}[t]
	\caption{Calculate Gradient Direction}
	\label{alg:getGradientDirection}
	\SetKwFunction{getGradientDirection}{}
	\SetKwProg{Fn}{getGradientDirection}{:}{}
	\Fn{\getGradientDirection{ $\normalfont \textbf{v}$ }}
	{
		$\widehat{V}_\textbf{v} \leftarrow \mathsf{getRandomSubset}(V_\textbf{v})$\;
		$\textbf{e} \leftarrow (1+\widehat{d}_{\textbf{v},\widehat{V}_\textbf{v}}) \frac{\partial}{\partial \textbf{v}} \mathrm{c}_\ell(\textbf{v}_\mathrm{p},\textbf{v}) 
+ \sum_{\textbf{u} \in \widehat{V_{\textbf{v}}}} (1+n_\textbf{u}) \frac{\partial}{\partial \textbf{v}} \mathrm{c}_\ell(\textbf{v},\textbf{u})$\;
        $\hat{\textbf{e}} \leftarrow \textbf{e} / \|\textbf{e}\|_2$
	}
	\KwRet $\hat{\textbf{e}}$
\end{algorithm}
\DecMargin{.5em}
\IncMargin{.5em}
\begin{algorithm}[t]
	\caption{Calculate Step-size}
	\label{alg:getStepSize}
	\SetKwFunction{getStepSize}{}
	\SetKwProg{Fn}{getStepSize}{:}{}
	\Fn{\getStepSize{ $\normalfont \textbf{v},\hat{\textbf{e}}$ }}
	{
		$\gamma_\mathrm{rel} \leftarrow \mathsf{getMaxStepSize}(\textbf{v},\hat{\textbf{e}})$\;
		$\gamma_\mathrm{max} \leftarrow \gamma_\mathrm{rel}$\;
		\While{$\gamma_\mathrm{max} > \delta $}
		{
			$u_\mathrm{rand}  \sim \mathcal{U}(0,1)$; 
			$\gamma \leftarrow (u_\mathrm{rand})^{1/d} \gamma_\mathrm{max}$\;
			$\textbf{x}_\mathrm{rand} \leftarrow  \textbf{v}-\gamma\hat{\textbf{e}}$\;
			\eIf{$\widehat{J}_{\mathcal{T},\widehat{V_\textbf{v}}}(\normalfont \textbf{x}_\mathrm{rand})<\widehat{J}_{\mathcal{T},\widehat{V_\textbf{v}}}(\normalfont \textbf{v})$}
			{
				$\mathsf{break}$\;
			}
			{
				$\gamma_\mathrm{max} \leftarrow \gamma $\;
			}
		}
		\If{$\gamma_\mathrm{max} < \delta$}
		{
		    $u_\mathrm{rand}  \sim \mathcal{U}(0,1)$; 
			$\gamma \leftarrow (u_\mathrm{rand})^{1/d} \gamma_\mathrm{rel}$\;
		}
	}
	\KwRet $\gamma$
\end{algorithm} 
\DecMargin{.5em}

This motivates the LES procedure, given in Algorithm~\ref{alg:LESalgFlow}. 
With probability $p_\mathrm{LES}$, LES is used to generate a new sample $\textbf{x}_\mathrm{rand}$ (Algorithm~\ref{alg:LESalgFlow}, line 6-8).
Otherwise, a new sample is generated using the conventional Informed Sampling technique given in \cite{gammell2018informed} (Algorithm~\ref{alg:LESalgFlow}, line 11).
This ensures a balance between exploration-exploitation (controlled by the parameter $p_\mathrm{LES}$) and graph growth in all the relevant homotopy classes. 
The $\mathsf{Extend}$ function takes this random sample and performs relevant procedures (nearest-neighbor, local steering and collision checking) to incorporate a new vertex in the graph (Algorithm~\ref{alg:LESalgFlow}, line 12). 
Finally, the graph-processing module operates on $\mathcal{G}$ considering the addition of a new vertex (Algorithm~\ref{alg:LESalgFlow}, line 13).

If the best solution cost $c_i$ after $i$ iterations is finite (indicating that a sub-optimal solution has been discovered), redundant exploration can be avoided by focusing the search on the Informed or Relevant Region set. 
As the Informed Set may be ineffective in focusing search for general cost-space problems, LES generates new samples in the Relevant Region $\mathcal{X}^{\epsilon}_\mathrm{rel}$ \cite{joshi2019relevant}, defined as follows 
\begin{equation}
\label{eq:relRegionSet}
    \mathcal{X}^{\epsilon}_\mathrm{rel} = \bigcup_{\textbf{v} \in V_\mathrm{rel}} \mathcal{B}^{\epsilon}_\mathrm{rel}(\textbf{v}),
\end{equation}
where,
\begin{equation}
\begin{aligned}
\mathcal{B}^{\epsilon}_\mathrm{rel}(\textbf{v}) & = \{\textbf{x} \in \mathcal{B}^{\epsilon}( \textbf{v}) \ | \ \hat{f}_\textbf{v}(\textbf{x}) < c_i \},\\
\hat{f}_\textbf{v}(\textbf{x}) & = \mathrm{c}_\ell(\textbf{v,x})+\mathrm{g}_\mathcal{T}(\textbf{v})+\mathrm{h}(\textbf{x},\textbf{x}_\mathrm{g}),
\end{aligned}
\end{equation}
and $V_\mathrm{rel}$ denotes the set of "relevant vertices", 
\begin{equation}
\label{eq:relVertices}
V_\mathrm{rel}= \{ \textbf{v} \in V \ | \ \mathrm{g}_\mathcal{T}(\textbf{v}) +\mathrm{h}(\textbf{v},\textbf{x}_\mathrm{g}) < c_i   \}.
\end{equation}  
The value of $\epsilon$ in (\ref{eq:relRegionSet}) is set to $\epsilon=1.5\eta$, where $\eta$ is the range parameter in SBMP algorithms \cite{sucan2012open}, which controls the maximum edge-length in $\mathcal{G}$.
The procedure for selecting a vertex ($\mathsf{chooseVertex}$), is similar to the implementation in \cite{joshi2019relevant}. 
It assigns a weight $q_\textbf{v}$ for each $\textbf{v} \in  V_\mathrm{rel}$ and uses a binary heap data-structure for sorting. 
Start, goal and leaf vertices (vertices with no children) are ignored by the $\mathsf{chooseVertex}$ function. 
\begin{figure*}
	\centering
	\includegraphics[width=0.64\columnwidth,height=0.35\columnwidth]{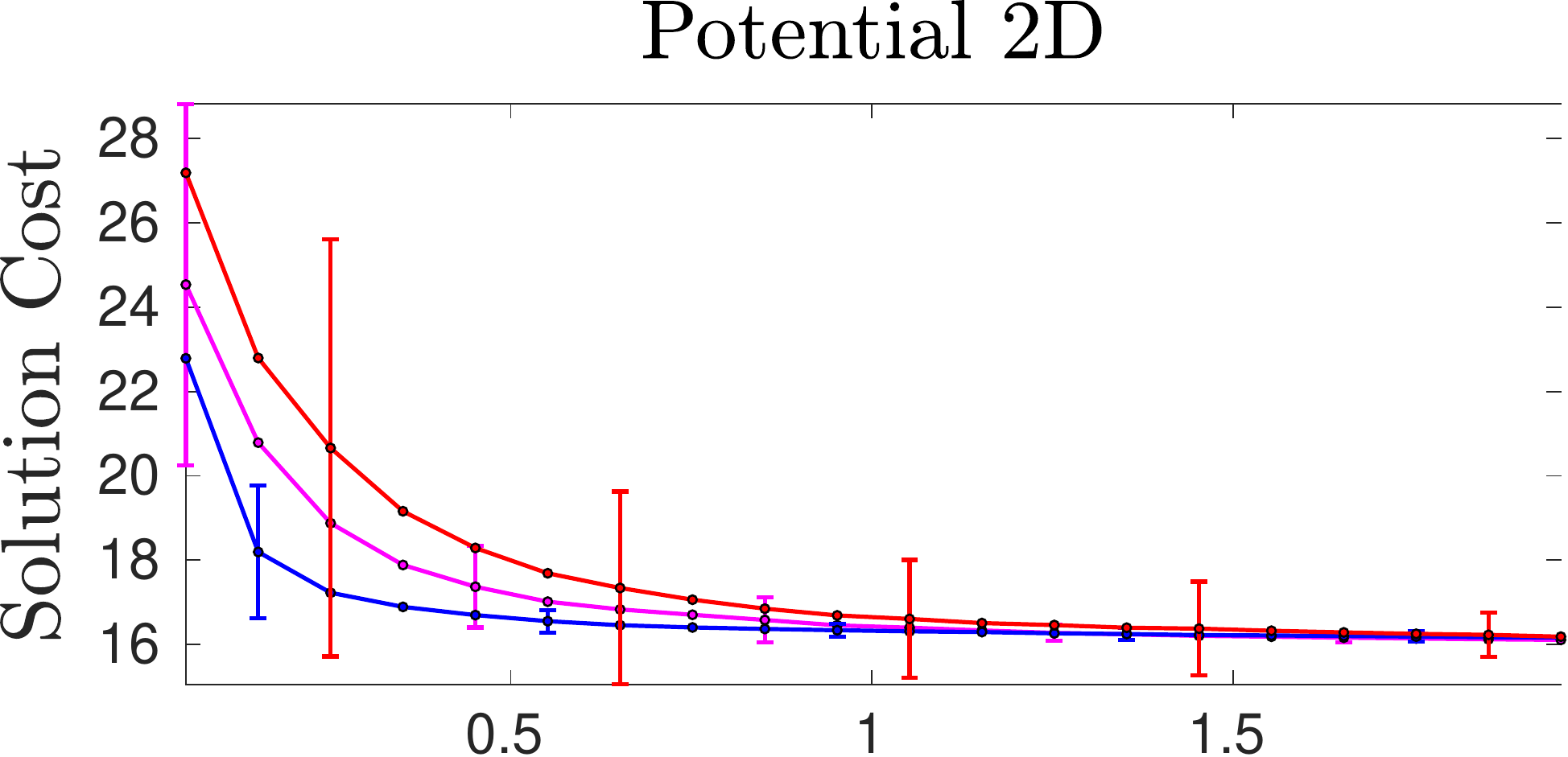}
	\includegraphics[width=0.64\columnwidth,height=0.35\columnwidth]{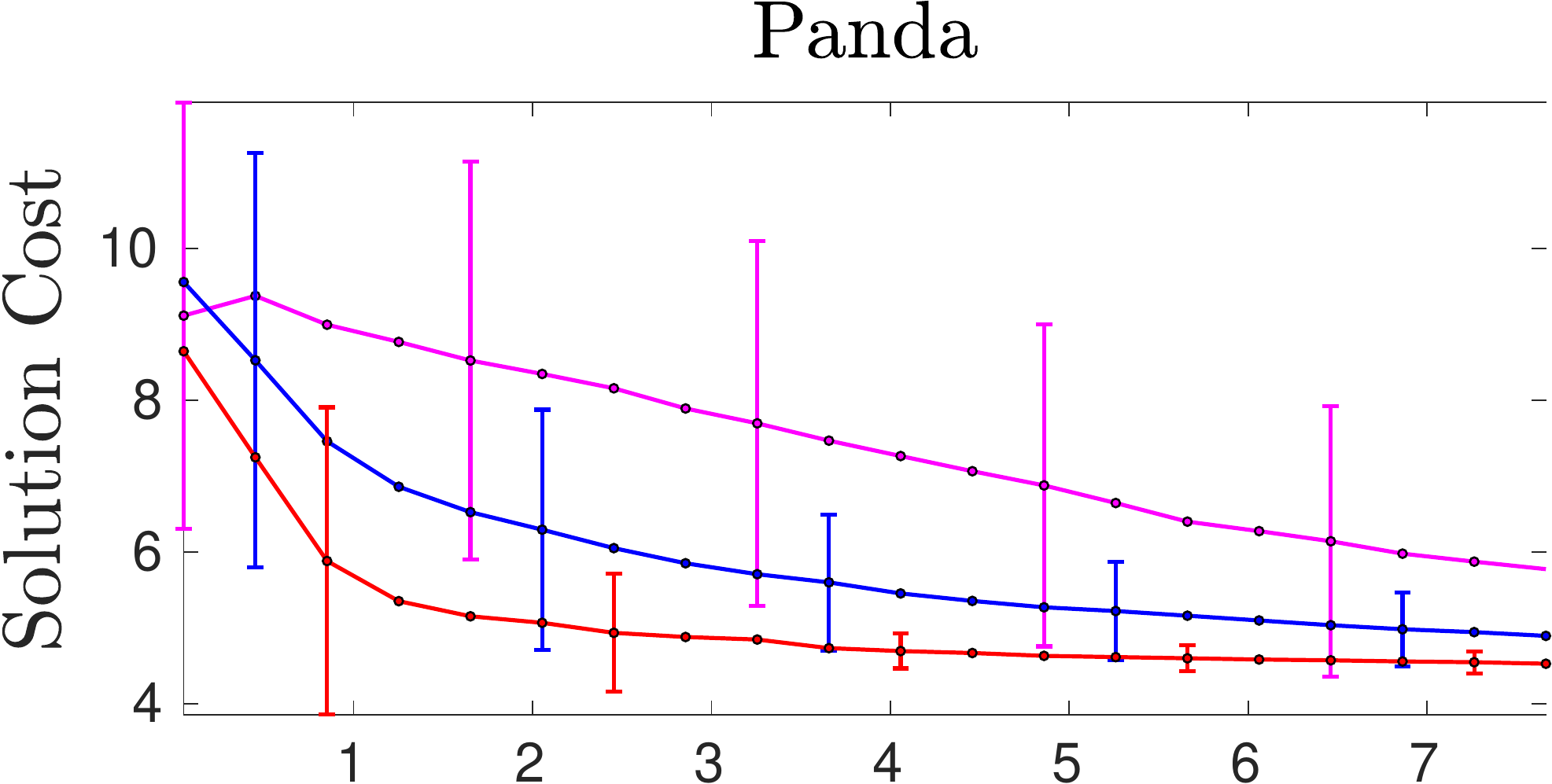}
	\includegraphics[width=0.64\columnwidth,height=0.35\columnwidth]{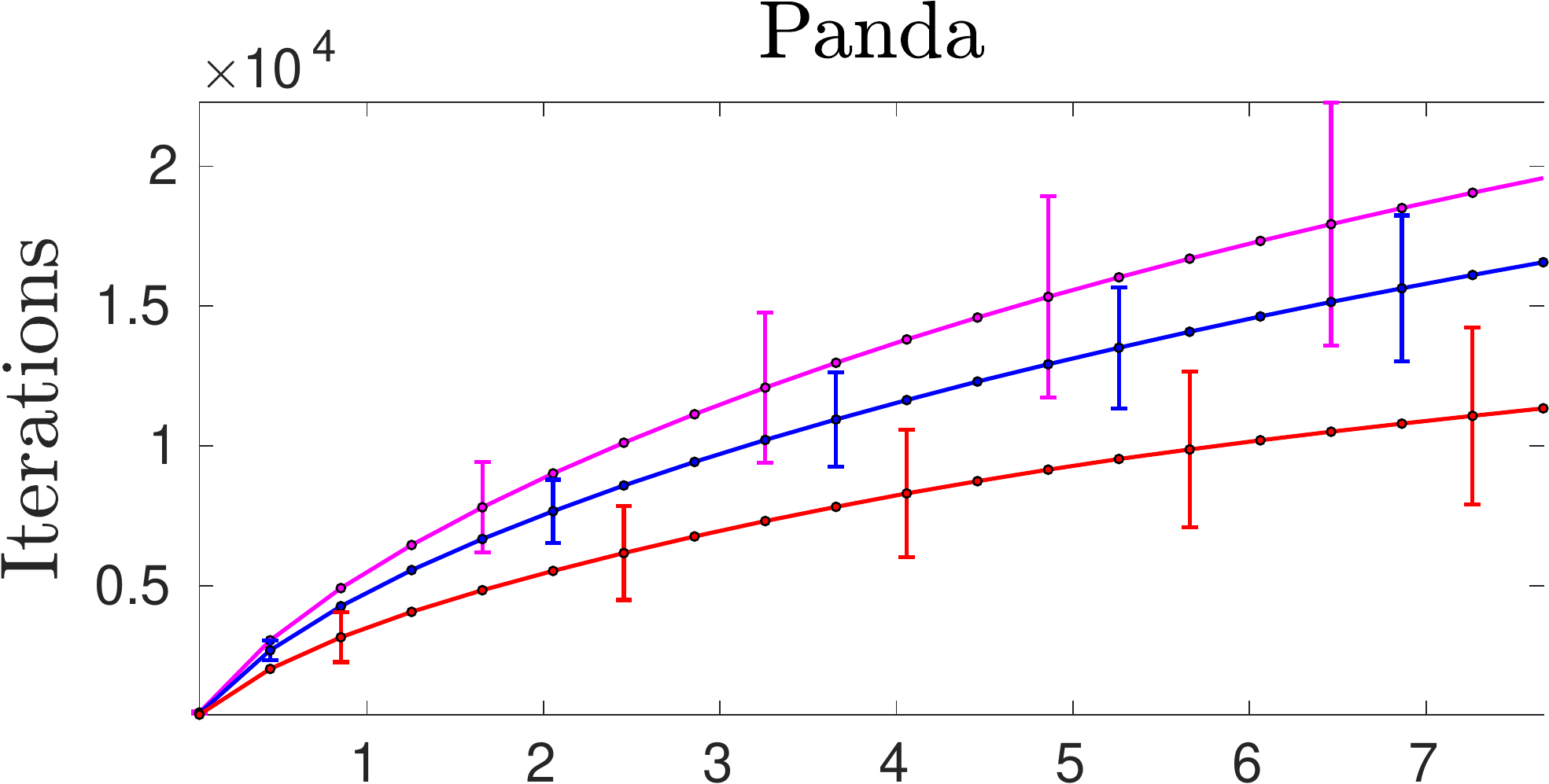}
	\vspace{0.5em}
	
	\includegraphics[width=0.64\columnwidth,height=0.35\columnwidth]{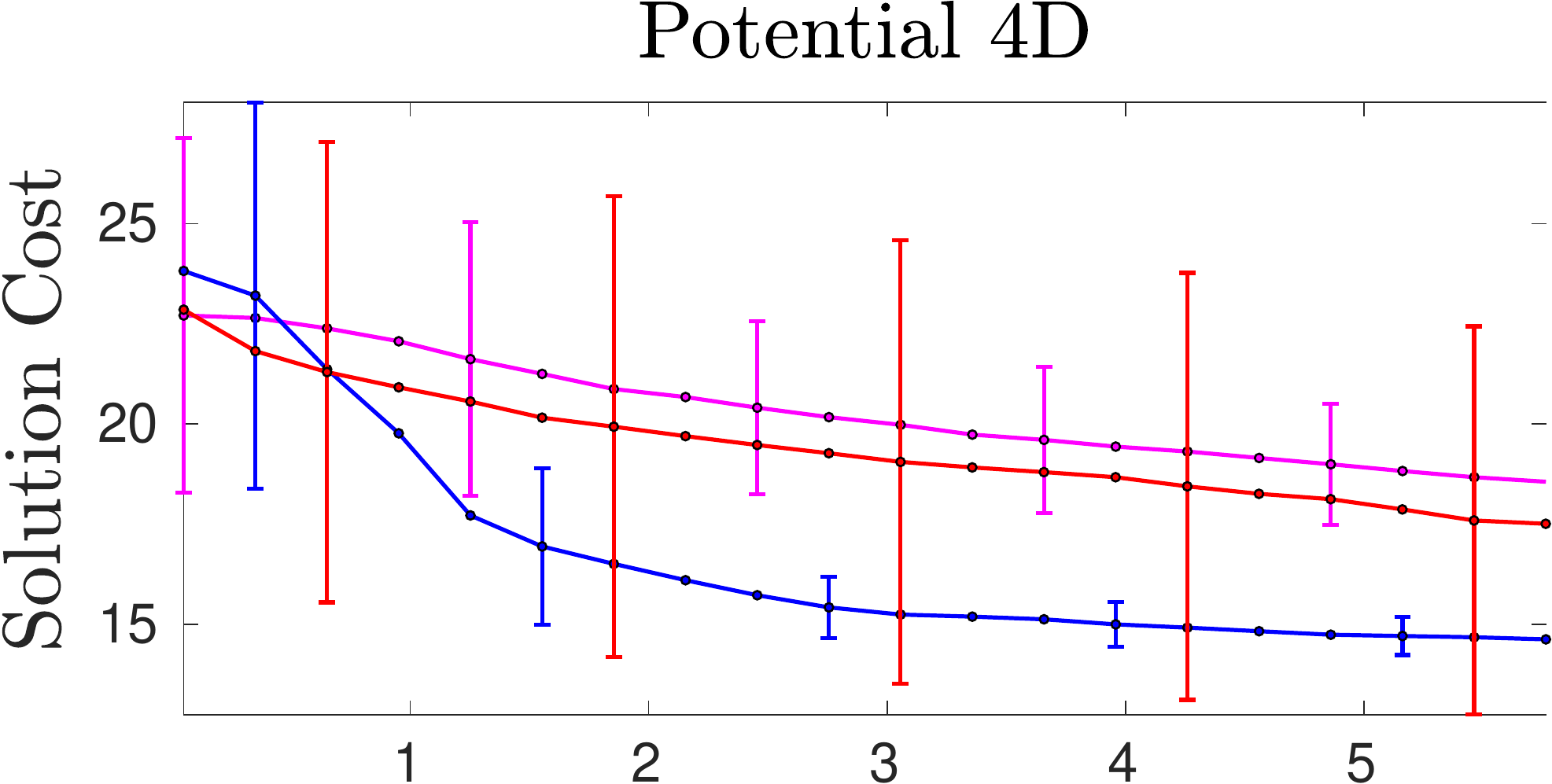}
	\includegraphics[width=0.64\columnwidth,height=0.35\columnwidth]{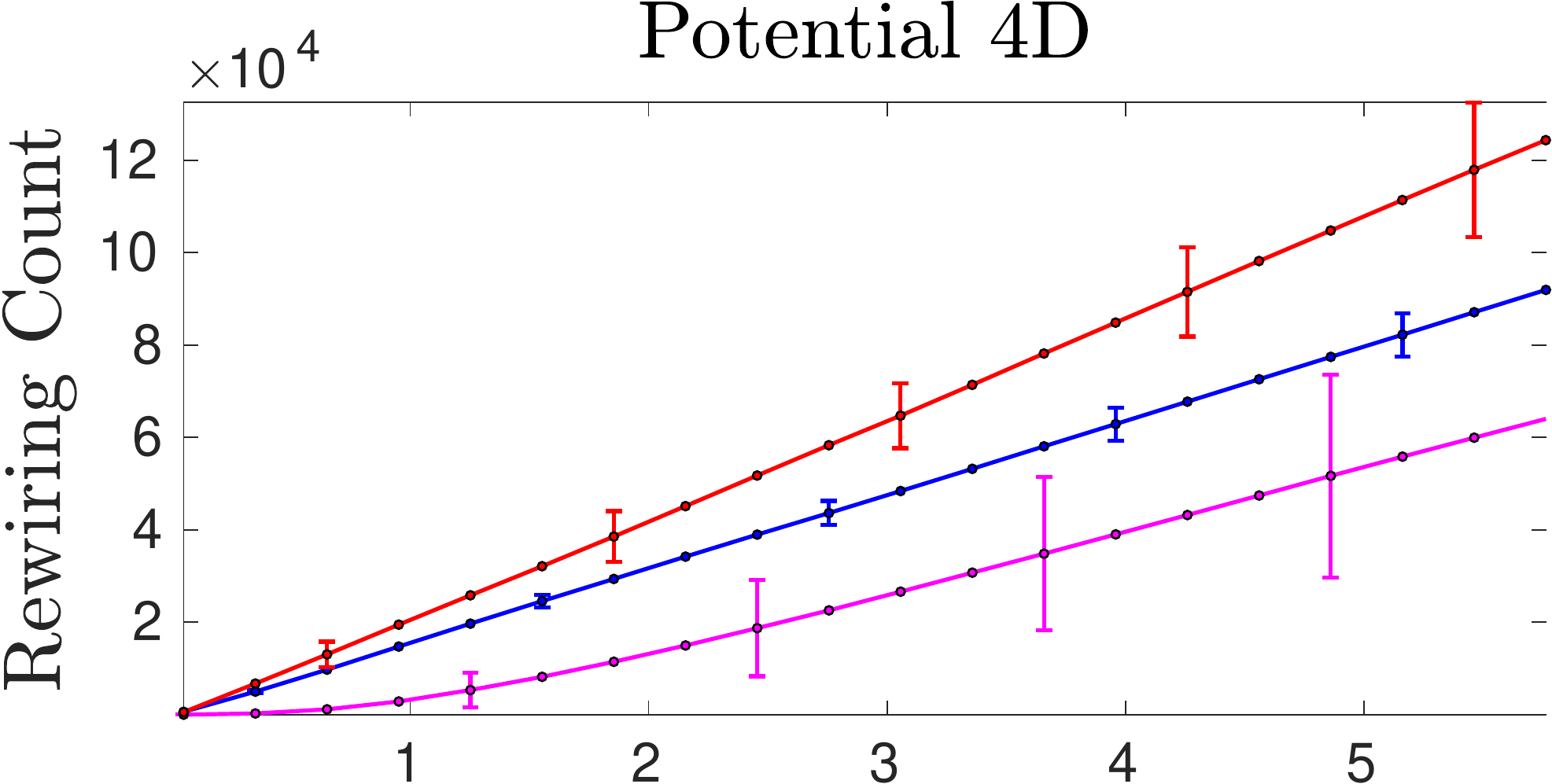}
	\includegraphics[width=0.64\columnwidth,height=0.35\columnwidth]{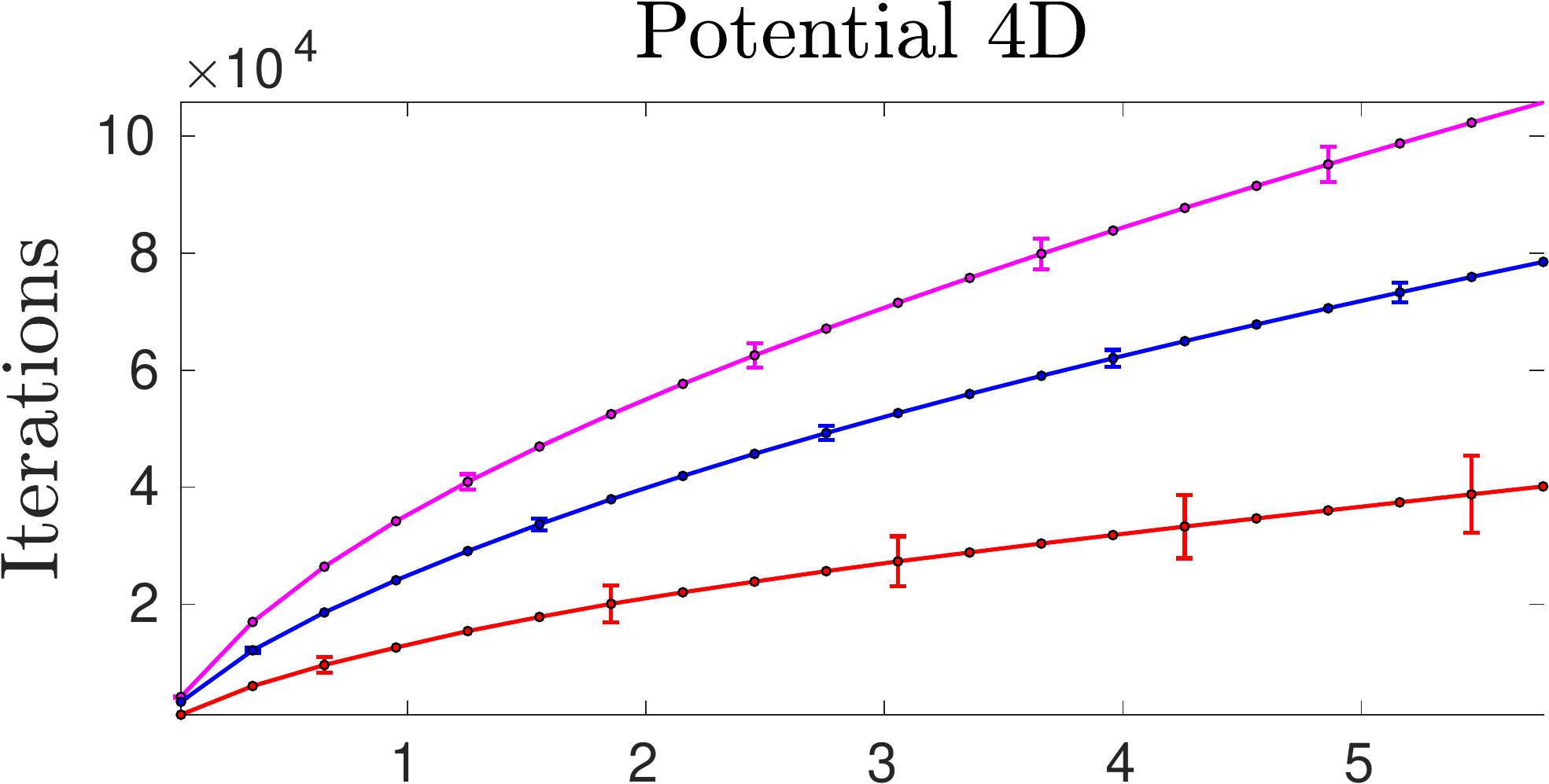}
	\vspace{0.5em}
	
	\includegraphics[width=0.64\columnwidth,height=0.35\columnwidth]{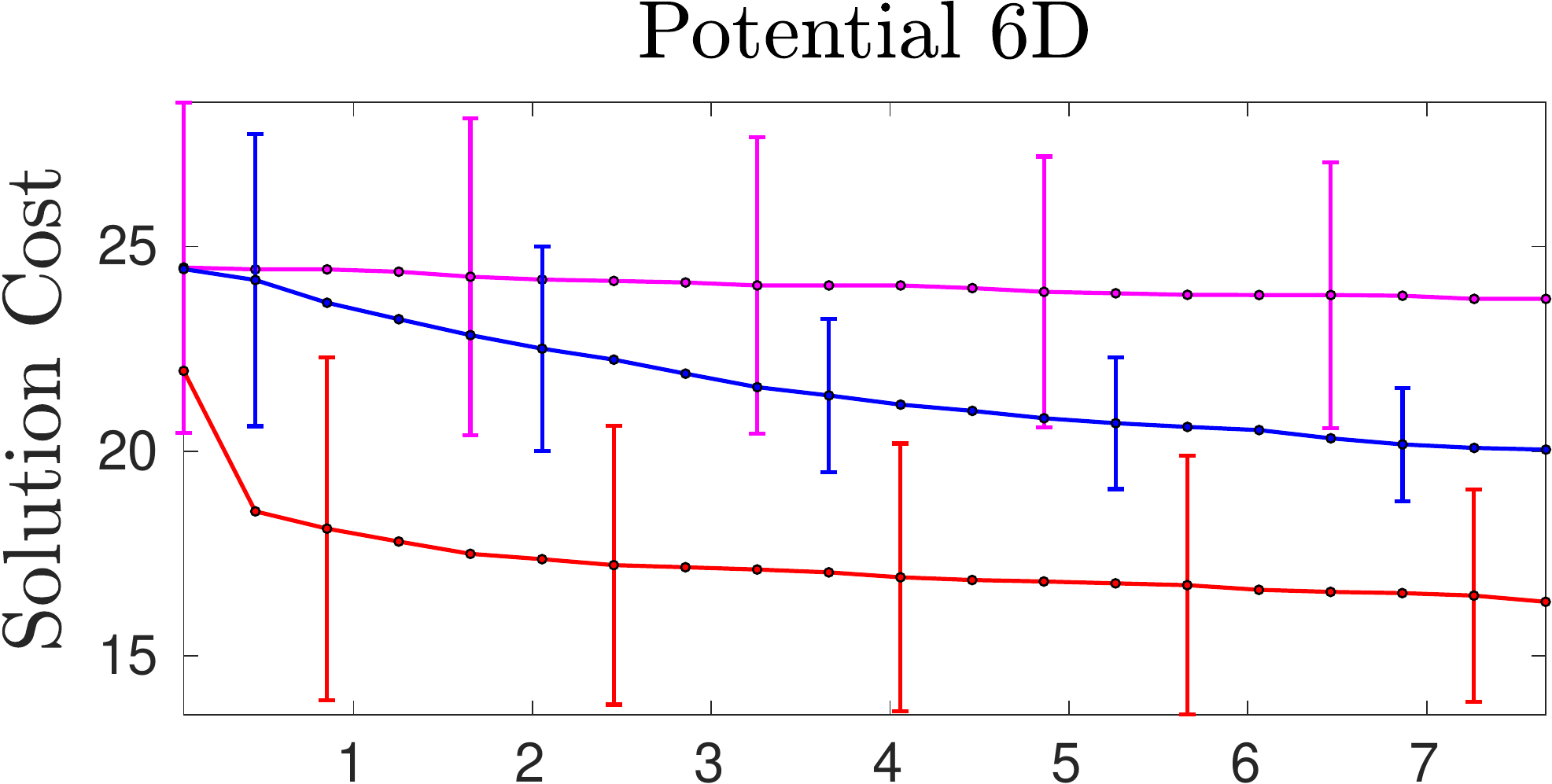}
	\includegraphics[width=0.64\columnwidth,height=0.35\columnwidth]{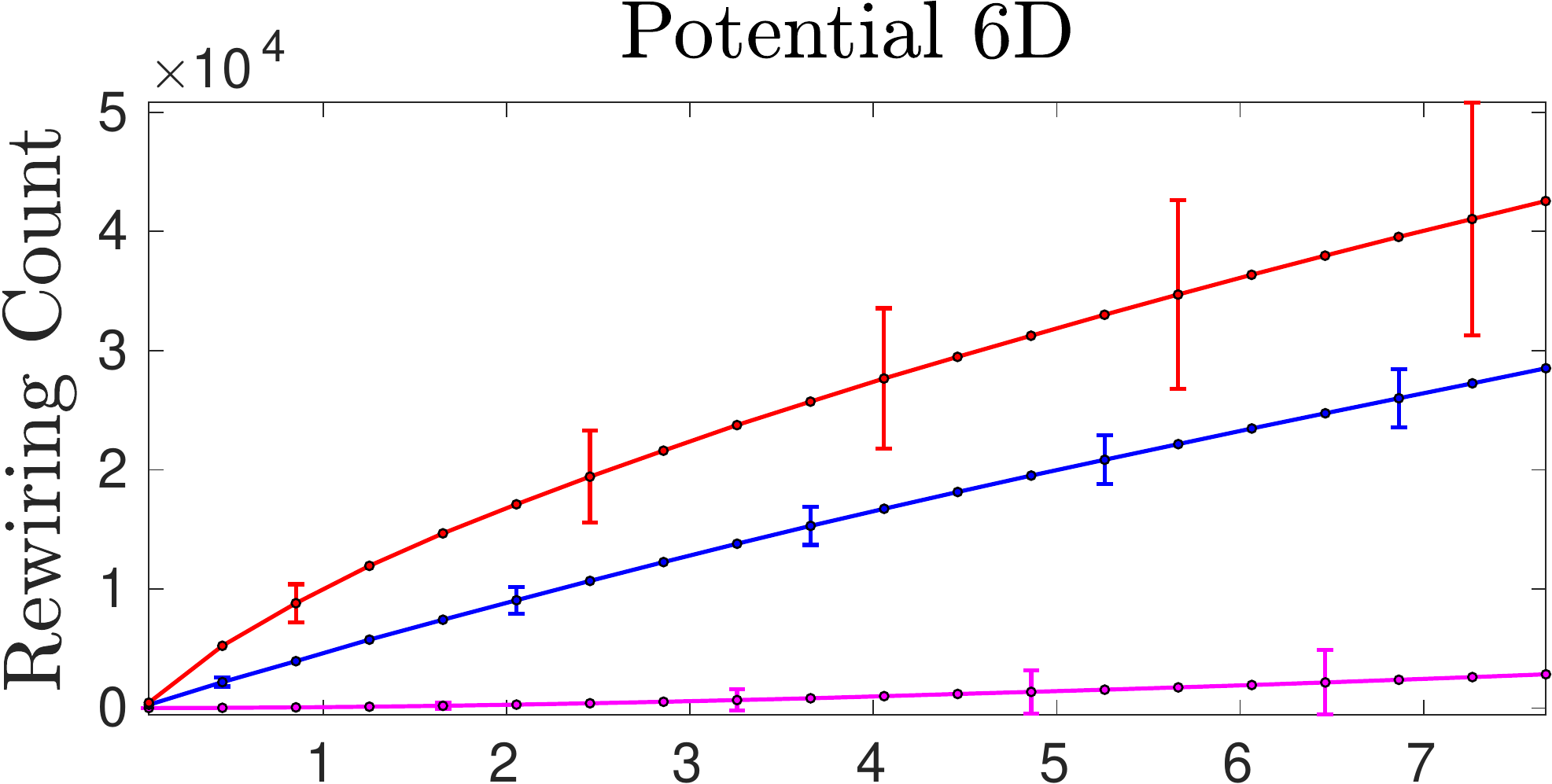}
	\includegraphics[width=0.64\columnwidth,height=0.35\columnwidth]{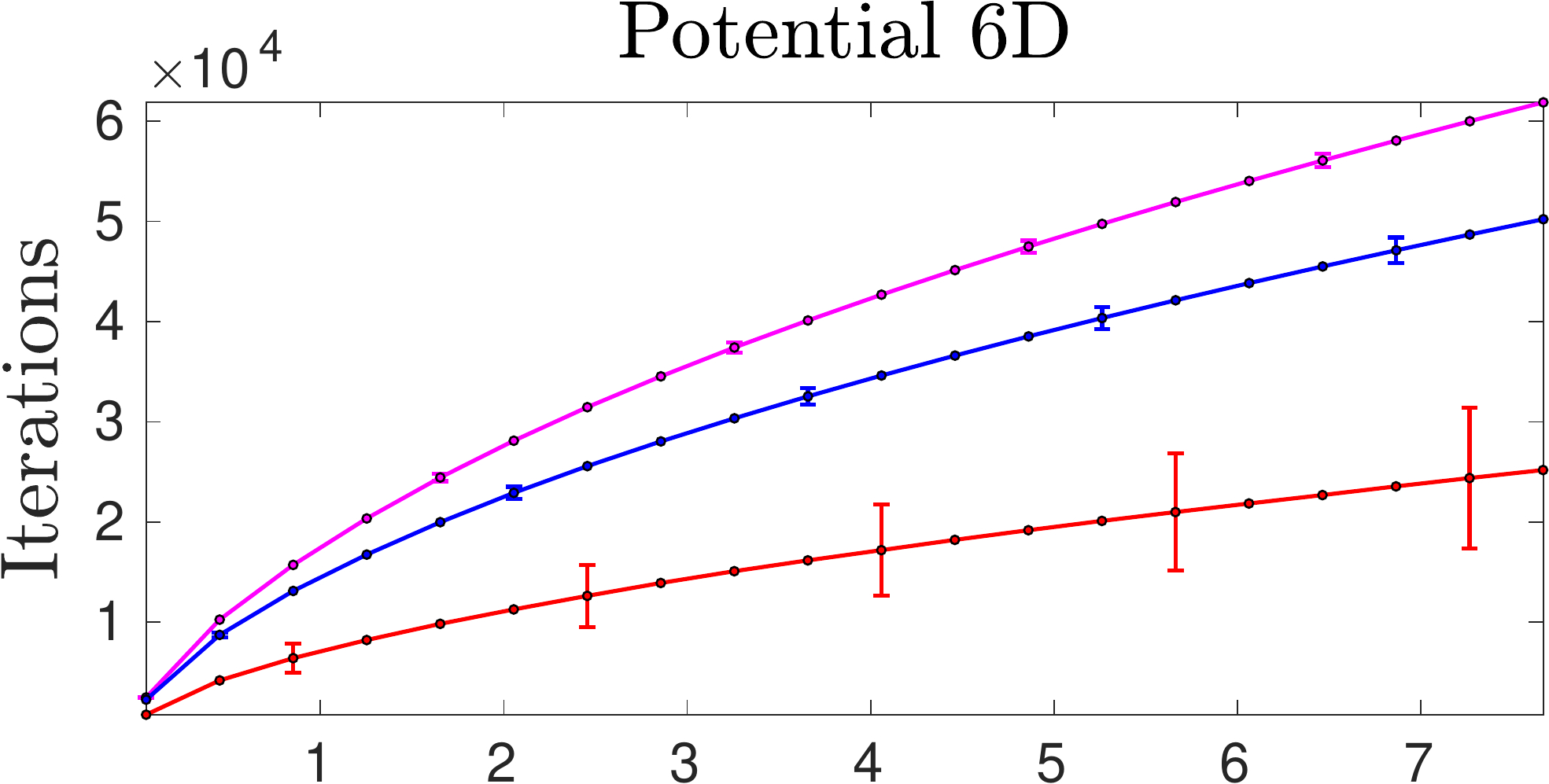}
	\vspace{0.5em}
	
	\includegraphics[width=0.64\columnwidth,height=0.35\columnwidth]{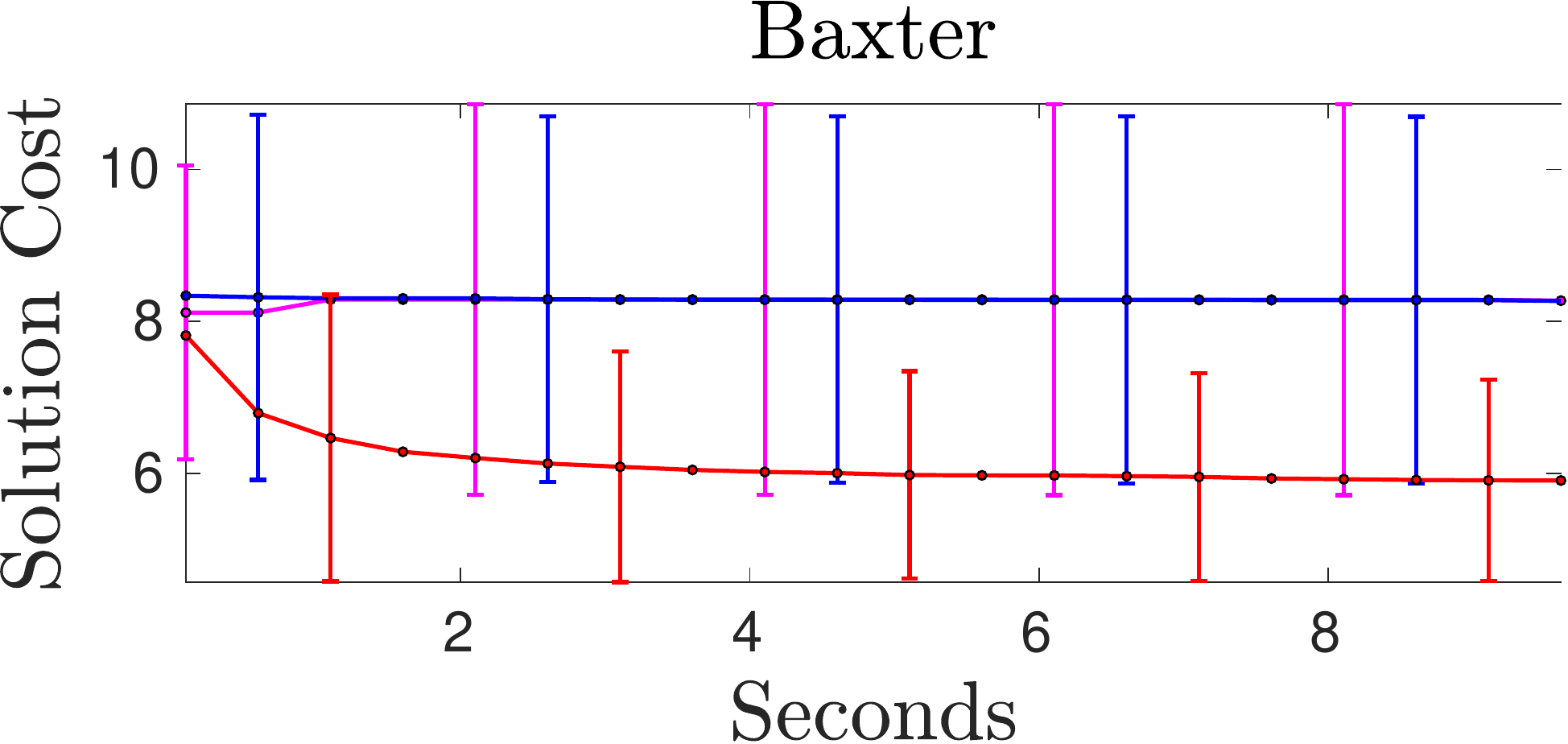}
	\includegraphics[width=0.64\columnwidth,height=0.35\columnwidth]{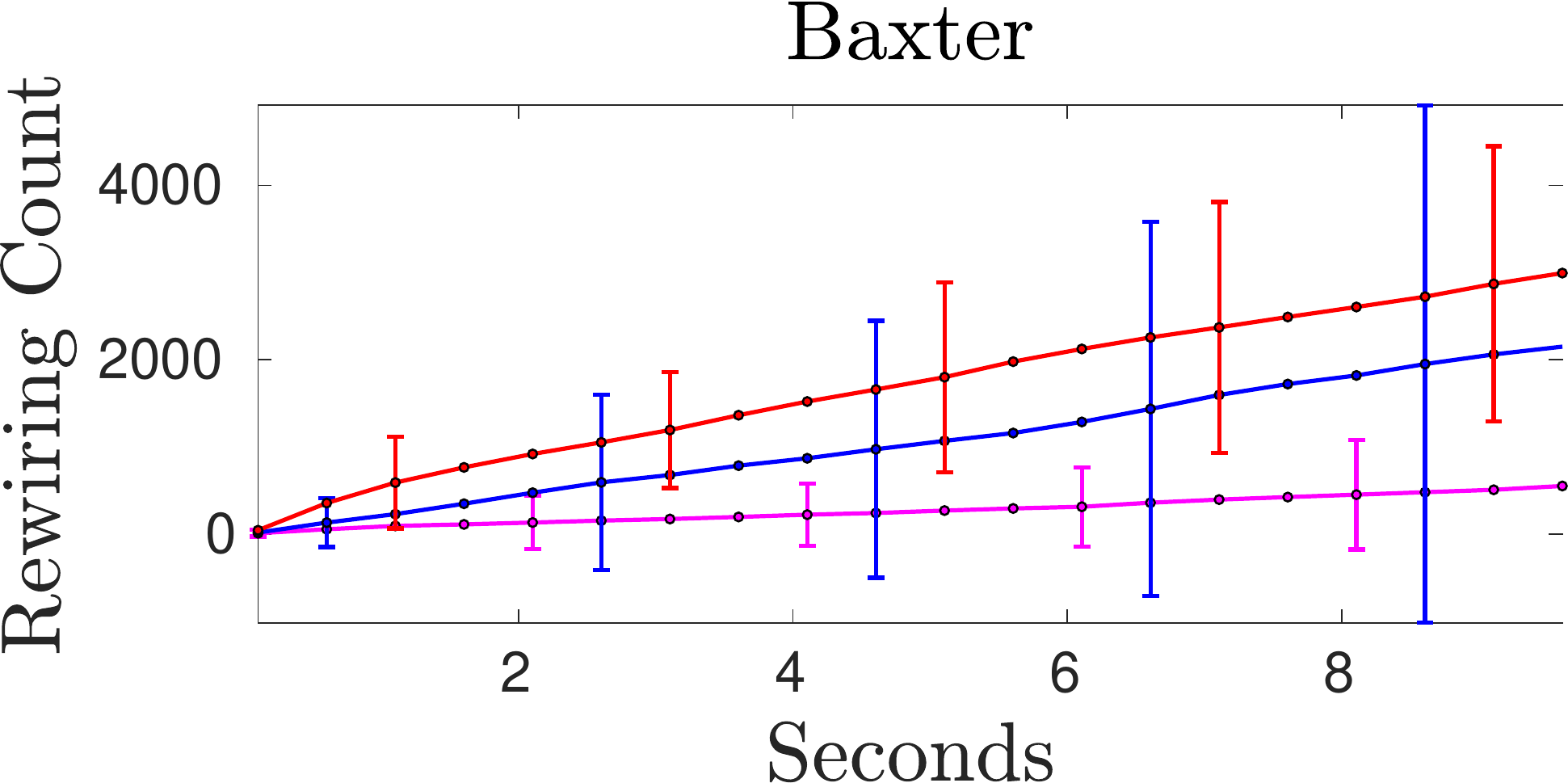}
	\includegraphics[width=0.64\columnwidth,height=0.35\columnwidth]{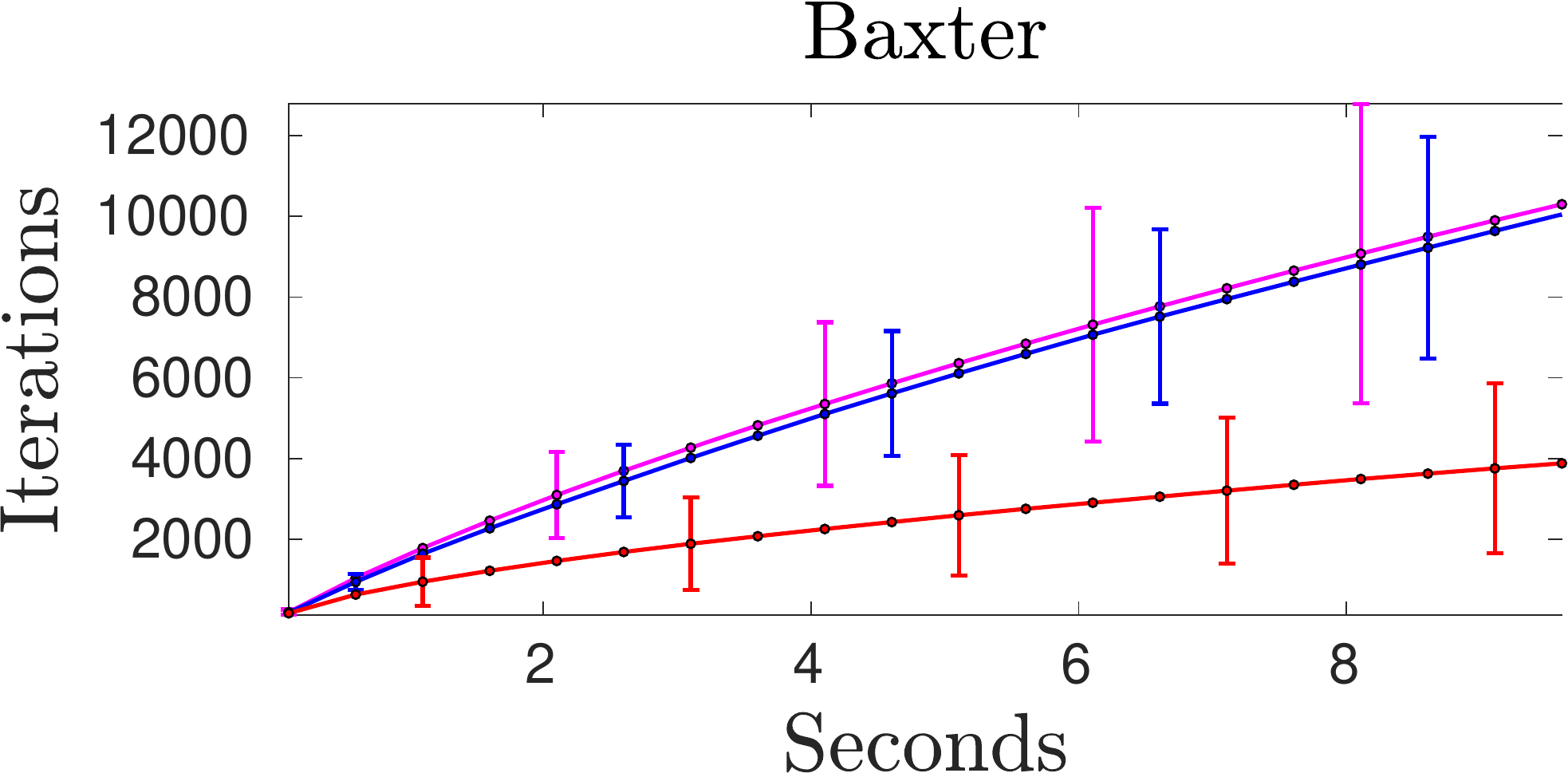}
	
	\includegraphics[width=0.8\columnwidth]{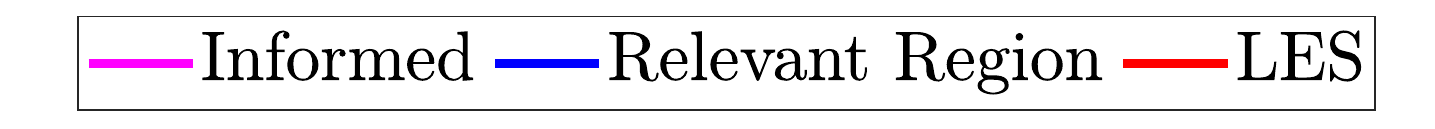}
	\caption{Benchmarking plots for the numerical experiments. Solid lines indicate the value averaged over 100 trials and the error bars represent standard deviation. Application of the proposed LES method (red) leads to a faster convergence and a larger number of tree rewirings in higher dimensions. However, it incurs a higher computational cost and hence executes a lesser number of iterations compared to Informed (magenta) and Relevant Region (blue) sampling.\vspace{-1em}}
	\label{fig:convergenceplots}
\end{figure*}

Note that $\widehat{J}_{\mathcal{T},\widehat{V_\textbf{v}}}(\textbf{v})$ represents a non-linear objective function. 
Hence, LES proceeds by numerically calculating the gradient of $\widehat{J}_{\mathcal{T},\widehat{V_\textbf{v}}}(\textbf{v})$ and moving an appropriate step-size in the direction of the gradient.
The procedure to calculate the gradient direction $\hat{\textbf{e}}$ is given in Algorithm~\ref{alg:getGradientDirection}.
First, a random subset of children $\widehat{V_\textbf{v}}$ is obtained. 
The gradient $\widehat{J}_{\mathcal{T},\widehat{V_\textbf{v}}}( \textbf{v})$ with respect to $\textbf{v}$ is calculated numerically using the symmetric difference formula (Algorithm~\ref{alg:getGradientDirection}, line 3).
Having obtained the gradient direction $\hat{\textbf{e}}$, the algorithm to calculate the step-size is given in Algorithm~\ref{alg:getStepSize}.
As finding the optimal step-size $\gamma^*$ by solving $\arg\min_{\gamma} \widehat{J}_{\mathcal{T},\widehat{V_\textbf{v}}}( \textbf{v}-\gamma\hat{\textbf{e}})$ is intractable, approaches such as backtracking line search \cite{boyd2004convex} have been suggested. 
However, executing backtracking line search is computationally not viable for the current application, as it requires a higher number of expensive calls to calculate $\widehat{J}_{\mathcal{T},\widehat{V_\textbf{v}}}$.
Instead, LES uses a procedure given in Algorithm~\ref{alg:getStepSize}, which is similar to the Hit-and-Run Sampler implemented in \cite{yi2018generalizing}. 
First, given a vertex $\textbf{v}$ and the travel direction $- \hat{\textbf{e}}$, the procedure in \cite{joshi2019relevant} is used to calculate the maximum step-size $\gamma_\mathrm{rel}$. 
This ensures that a candidate $\textbf{v}-\gamma\hat{\textbf{e}} \in \mathcal{X}^{\epsilon}_\mathrm{rel}$ for any $\gamma \in (0,\gamma_\mathrm{rel})$.
Variable $\gamma_\mathrm{max}$ is set to $\gamma_\mathrm{rel}$.
Next, a random step-size $\gamma$ is sampled from the interval $(0,\gamma_\mathrm{max})$.
The exponent of $1/d$ in Algorithm~\ref{alg:getStepSize}, line 5 biases $\gamma$ towards $\gamma_\mathrm{max}$.
If the candidate $\textbf{v}-\gamma\hat{\textbf{e}}$ results in an improvement for $ \widehat{J}_{\mathcal{T},\widehat{V_\textbf{v}}}$, step-size $\gamma$ is returned. 
Else, $\gamma_\mathrm{max}$ is updated to $\gamma$.
Thus, the search interval is sequentially reduced until a suitable step-size is discovered. 
Theoretically, a travel of infinitesimal magnitude in the direction of the gradient always results in an improvement. 
However, if $\gamma_\mathrm{max}$ is less than a small quantity $\delta << \eta$, then a random $\gamma$ in the interval $(0,\gamma_\mathrm{rel})$ is returned (Algorithm.~\ref{alg:getStepSize}, line 11-12) to avoid clumping of new vertices around $\textbf{v}$.
\section{NUMERICAL EXPERIMENTS}
The proposed LES algorithm was benchmarked against Informed sampler and Relevant Region sampler described in \cite{gammell2018informed} and \cite{joshi2019relevant} respectively.
Note that LES and Relevant Region sampler share a similar $\mathsf{chooseVertex}$ procedure.
However, the Relevant Region sampler only generates random samples in $\mathcal{X}^{\epsilon}_\mathrm{rel}$ and does not consider the optimization problem corresponding to (\ref{eq:JTv_hat}).
All the algorithms were implemented using C++/OMPL \cite{sucan2012open}. 
Data was gathered over 100 trials for each experiment using the standardized OMPL benchmarking tools \cite{moll2015benchmarking}.
All experiments were performed on a 64 bit laptop running Ubuntu 16.04 OS, with 16 GB RAM and an Intel i7 processor. 
The parameter $p_\mathrm{LES}$ and an analogous parameter $p_\mathrm{rel}$ for Relevant Region sampler were both set to $0.5$.
Parameter $\delta$ was set to $10^{-4}$.
All sampling strategies used a goal bias of $5\%$ and were paired with RRT$^{\#}$'s global rewiring for graph-processing.
A description of the different benchmarking environments is given below.

\noindent \textbf{Potential Cost-map:} This environment, illustrated in Fig.~\ref{fig:potential_costmap}, has the state-cost function
\begin{equation}
    C(\textbf{x})=1 + 9\sum_{i} \exp{ \big( -\|\textbf{x}^\mathrm{c}_i -\textbf{x} \|_2^2 \big) }.
\end{equation}
Here, $\textbf{x}^\mathrm{c}_i$ represent the center points of the high cost white regions. The objective for the robot is to plan a path to the goal while avoiding these soft obstacles. 
The range parameter $\eta$ was set to $0.4, 0.6$ and $1.5$ for the 2D, 4D and 6D versions of environment respectively. 

\noindent \textbf{Robot Manipulators:}
A planning problem for a 7 DOF Panda and a 14 DOF Baxter arm is illustrated in Fig.~\ref{fig:manipulators}. 
The objective was to find the minimum length path ($ C(\textbf{x})=1$ for all $\textbf{x} \in \mathcal{X}$) in the configuration-space with strict joint limits ($\mathbb{R}^7$ for Panda, $\mathbb{R}^{14}$ for Baxter). 
These joint limits and collision checking calculations were implemented with the help of MoveIt!~\cite{chitta2012moveit}.
The range parameter $\eta$ was set to $1.2$ and $2$ for the Panda and Baxter experiments respectively.

Results from the numerical experiments are illustrated in Fig.~\ref{fig:convergenceplots}. 
The proposed LES algorithm outperforms Informed (magenta) and Relevant Region (blue) samplers in higher dimensional settings (Potential 6D, Panda, Baxter) in terms of cost convergence. 
LES also initiates a larger number of rewirings in $\mathcal{T}$.
However, similar performance gains are not seen in the lower dimensional environments (Potential 2D, 4D).
Relevant Region sampler, with its focusing properties performs better than Informed sampling. 
LES incurs a higher computational cost due to the numerical gradient calculations in Algorithm~\ref{alg:getGradientDirection} and expensive function evaluations of $\widehat{J}_{\mathcal{T},\widehat{V_\textbf{v}}}$ in Algorithm~\ref{alg:getStepSize}.
Thus, the application of LES leads to a lesser number of iterations executed in a given time period compared to the other two methods. This might slow down convergence in lower dimensions. 
However, random search techniques are affected by the "curse of dimensionality" as illustrated in the Appendix.
This justifies the computationally costly procedures of LES which lead to an accelerated convergence in higher dimensions.

\section{CONCLUSION}
This work proposes a "Locally Exploitative Sampling" algorithm, that generates new samples to improve the cost-to-come value of vertices in a neighborhood. 
LES numerically calculates the gradient of (\ref{eq:JTv_hat}) and decides an appropriate step-size to obtain a new sample. 
Although computationally costlier, LES adds an "exploitative-bias" that can accelerate convergence of SBMP algorithms, especially in higher dimensions. 
LES generates new samples in the Relevant Region, a subset of the Informed Set, to avoid redundant exploration after an initial solution is discovered.
As discussed earlier, Informed Sampling is a necessary condition to improve the current solution. However, it is not sufficient, as an "Informed sample" is not guaranteed to bring about improvements in the current solution or the cost-to-come value of vertices. 
LES can be seen as a way to address this limitation of Informed Sampling. 

LES presents many openings for future research. 
LES can be extended to kino-dynamic settings and be used with planners such SST~\cite{li2015sparse}.
While the current implementation does not leverage the obstacle data gathered by the planner, ideas from \cite{lai2020bayesian} can be used to have a "obstacle-aware" version of LES. 
Exploring depth-$k$ generalization of the objective (\ref{eq:JTv_hat}) and analysing its effect on convergence and computational cost is also the focus of future work.
\begin{figure}
    \centering
    \includegraphics[width=0.7\columnwidth]{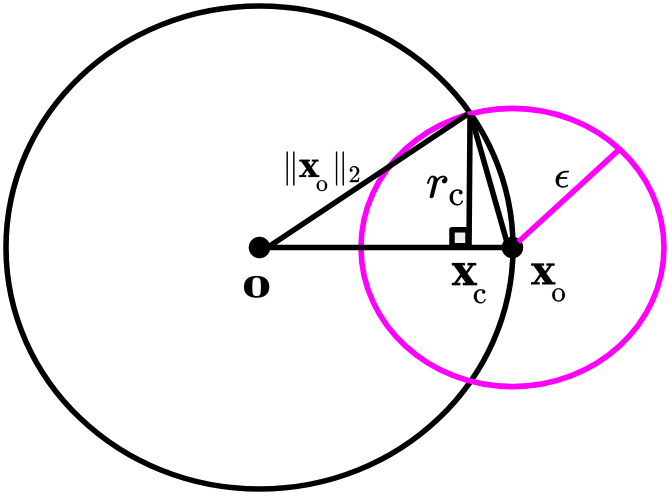}
    \caption{Schematic for the analysis in Appendix. Black and magenta circles illustrate the set $\mathcal{B}^{\|\textbf{x}_\mathrm{o}\|_2}( \textbf{0})$ and $\mathcal{B}^\epsilon(\textbf{x}_\mathrm{o})$ respectively. The intersection $\mathcal{B}^{\epsilon}( \textbf{x}_\mathrm{o}) \cap \mathcal{B}^{\|\textbf{x}_\mathrm{o}\|_2}( \textbf{0})$ can be over-approximated by hyper-sphere centered at $\textbf{x}_\mathrm{c}$ with radius $r_c$.}
    \label{fig:cod_motivate}
\end{figure}
\section*{APPENDIX}
\label{sec:appendix}
The following analysis is similar to the one provided in \cite{watt2016machine}.
Consider the problem of minimizing a quadratic objective function $J_q(\textbf{x})=\textbf{x}^\mathsf{T}\textbf{x}$ with random local search.
Let the starting state be $\textbf{x}_\mathrm{o} \in \mathbb{R}^d$ with the corresponding objective cost $J_q(\textbf{x}_\mathrm{o})$.
Random search generates samples in the set $\mathcal{B}^\epsilon(\textbf{x}_\mathrm{o})$ to find a new state with cost less than $J_q(\textbf{x}_\mathrm{o})$. 
Assume $\epsilon<\|\textbf{x}_\mathrm{o}\|_2$.
The set of states that provide an improvement over $J_q(\textbf{x}_\mathrm{o})$ satisfy  $\textbf{x}^\mathsf{T}\textbf{x}<\textbf{x}_\mathrm{o}^\mathsf{T}\textbf{x}_\mathrm{o}$. 
This set can be denoted as $\mathcal{B}^{\|\textbf{x}_\mathrm{o}\|_2}( \textbf{0})$, where $\textbf{0}$ is the origin.
The set of good samples thus lie in the set $\mathcal{B}^{\epsilon}( \textbf{x}_\mathrm{o}) \cap \mathcal{B}^{\|\textbf{x}_\mathrm{o}\|_2}( \textbf{0})$.
Please see Fig.~\ref{fig:cod_motivate}.
This intersection between two hyper-spheres can be over-approximated by $\mathcal{B}^{r_c}(\textbf{x}_\mathrm{c})$, where
\begin{equation}
    \label{eq:r_c}
    r_c = \epsilon\sqrt{1-\frac{\epsilon^2}{4\|\textbf{x}_\mathrm{o}\|^2_2}}.
\end{equation}
The probability of generating a good sample using random search is given by 
\begin{equation}
\begin{aligned}
    \mathbb{P}(\textbf{x} \in \mathcal{B}^{\epsilon}( \textbf{x}_\mathrm{o}) \cap \mathcal{B}^{\|\textbf{x}_\mathrm{o}\|_2 }( \textbf{0}))
    & = \frac{\mu \big( \mathcal{B}^{\epsilon}( \textbf{x}_\mathrm{o}) \cap \mathcal{B}^{\|\textbf{x}_\mathrm{o}\|_2 }( \textbf{0}) \big) }
    {\mu(\mathcal{B}^{\epsilon}( \textbf{x}_\mathrm{o}))}\\
    & < \frac{\mu(\mathcal{B}^{r_c}(\textbf{x}_\mathrm{c}))}
    {\mu(\mathcal{B}^{\epsilon}( \textbf{x}_\mathrm{o}))}\\
    & = \big( 1-\frac{\epsilon^2}{4\|\textbf{x}_\mathrm{o}\|^2_2} \big)^{\frac{d}{2}}.
\end{aligned}
\end{equation}
Thus, the probability of generating a good sample decreases exponentially with the dimension $d$.

\noindent \textbf{Acknowledgements}:
This work has been supported by NSF awards IIS-1617630 and IIS-2008686.
\bibliographystyle{IEEEtran}
\bibliography{references}	

\end{document}